%% file: 0_main.tex
\documentclass[letterpaper, 10 pt, conference]{ieeeconf}  

\IEEEoverridecommandlockouts                              

\usepackage{graphicx} 
\usepackage{amsmath} 
\usepackage{amssymb}  
\usepackage{booktabs}
\usepackage{soul}

\usepackage{algorithm}
\usepackage{algpseudocode}
\usepackage{graphicx}
\usepackage{xcolor}

\usepackage{amsthm}
\usepackage{hyperref}
\usepackage{tabularx}

\newcommand{\eg}{\textit{e.g.}}

\title{\LARGE \bf
    Reactive Human-to-Robot Handovers of Arbitrary Objects
}

\usepackage[font=scriptsize,labelfont=bf]{caption}
\author{Wei Yang$^{\dagger}$, Chris Paxton$^{\dagger}$, Arsalan Mousavian$^{\dagger}$, Yu-Wei Chao$^{\dagger}$, Maya Cakmak$^{\ddagger}$, Dieter Fox$^{\dagger,\ddagger}$
    \thanks{$^{\dagger}$NVIDIA, USA
            {\tt\small \{weiy, cpaxton, amousavian, ychao, dieterf\}@nvidia.com}}%
    \thanks{$^{\ddagger}$University of Washington, USA
            {\tt\small mcakmak@cs.washington.edu}}%
}

\begin{document}

\maketitle
\thispagestyle{empty}
\pagestyle{empty}

\begin{abstract}
\input{1_abstract}
\end{abstract}

\section{Introduction}\label{section:intro}
\input{2_introduction}

\section{Related Work}\label{section:related_work}
\input{3_related_work}

\section{Perception for Arbitrary Object Handovers}\label{section:hand_model}
\input{4_perception}

\section{Grasp Selection for Reactive Handovers}\label{section:task_model}
\input{5_planning}

\section{Systematic Evaluation}\label{section:sys_eval}

\input{6_experiments}

\section{User Study with Household Objects}\label{section:user_study}
\input{7_user_study}

\section{Conclusion} 
\label{section:conclusion}
\input{8_discussion}






\bibliographystyle{IEEEtran}
\bibliography{handover}

\clearpage
\input{appendix}

\end{document}

%% file: 1_abstract.tex
Human-robot object handovers have been an actively studied area of robotics over the past decade; however, very few techniques and systems have addressed the challenge of handing over diverse objects with arbitrary appearance, size, shape, and deformability. 
In this paper, we present a vision-based system that enables reactive human-to-robot handovers of unknown objects. Our approach combines closed-loop motion planning with  real-time, temporally consistent grasp generation to ensure reactivity and motion smoothness.
Our system is robust to different object positions and orientations, and can grasp both rigid and non-rigid objects.
We demonstrate the generalizability, usability, and robustness of our approach on a novel benchmark set of $26$ diverse household objects, a user study with six participants  
handing over a subset of $15$ objects, and a systematic evaluation examining different ways of handing objects.

%% file: 2_introduction.tex
The ability to handover objects to and from humans is critical for robots that are intended to assist people in different environments. Robots can increase efficiency in factories by handing parts and tools to workers, or bring independence to older adults and people with mobility limitations at home by fetching items needed for daily activities. A growing community of robotics researchers have tried to address challenges in enabling seamless autonomous handovers, from human and object perception to motion control and communication during handovers \cite{ortenzi2020object}.

One of the key challenges of enabling generalizable and robust handover behaviors on robots is the diversity of objects that they need to handle, especially in unstructured, personalized environments like homes. 
For example, Choi et al.~\cite{choi2009list} conducted a study with ALS patients to identify the objects that they needed most on a daily basis by their caregivers or service dogs, to inform the design of a robot for the same purpose. The list includes over 40 objects, from remote control and utensils to pillows and newspapers, covering a wide range of sizes, shapes, appearances, and deformability. In addition to the variations across different categories of objects, even one type of object can vary drastically from one home to another in these dimensions. 
Despite the great progress made on human-robot handover research, the diversity of objects with which existing methods have been evaluated is far less than what is needed to address this challenge. Our work aims to close this gap.

\begin{figure}
    \centering
    \includegraphics[width=1\linewidth]{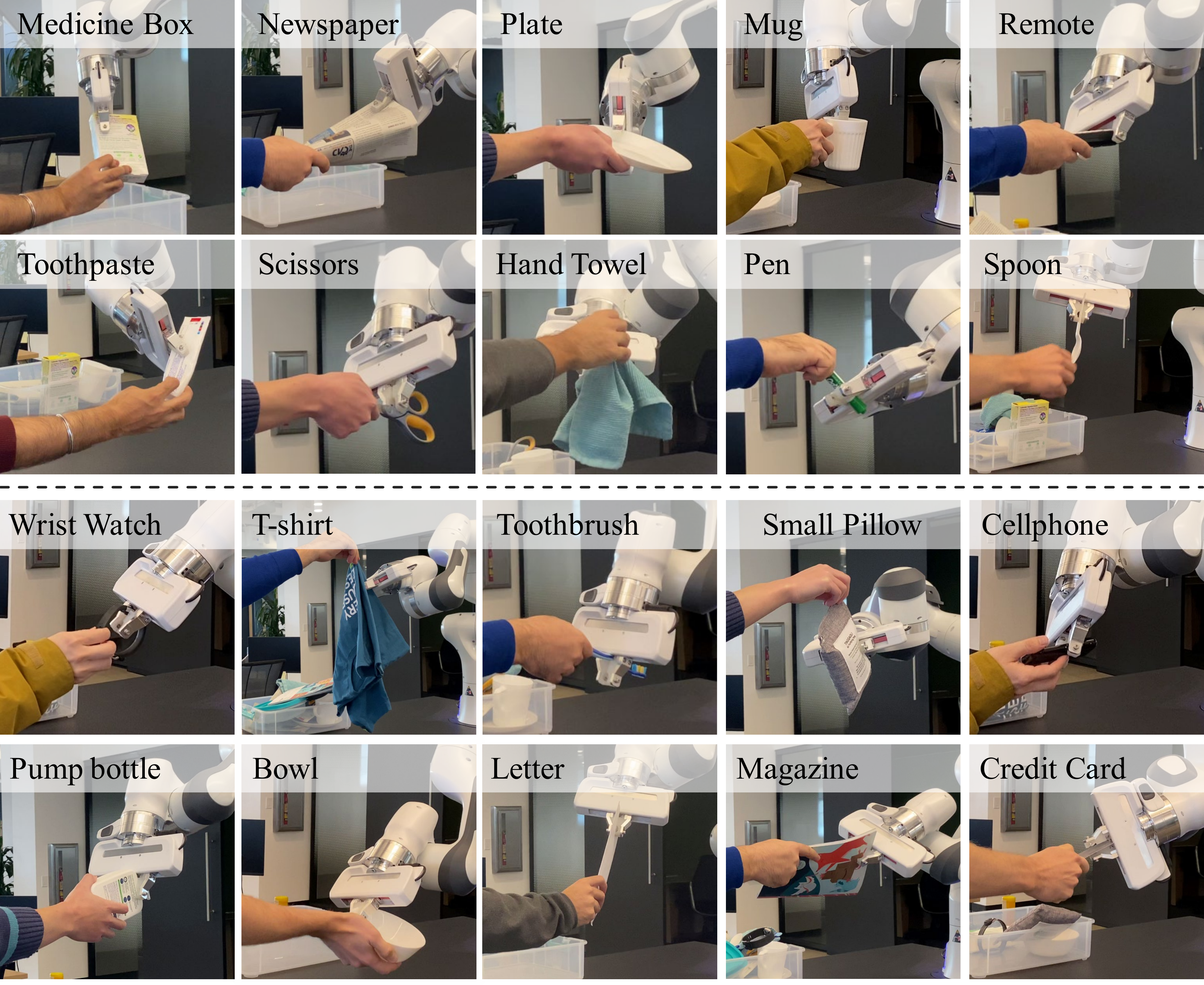}
    \caption{Our approach allows human-to-robot handovers of arbitrary objects that are graspable. It was evaluated with a benchmark set of household objects from Choi et al.'s work~\cite{choi2009list} that identifies a list of objects most needed in assistive settings. Videos are available at \url{https://sites.google.com/nvidia.com/handovers-of-arbitrary-objects}.
    }
    \label{fig:teaser}
\end{figure}

\begin{figure*}[ht]
    \centering
    \includegraphics[width=1\textwidth]{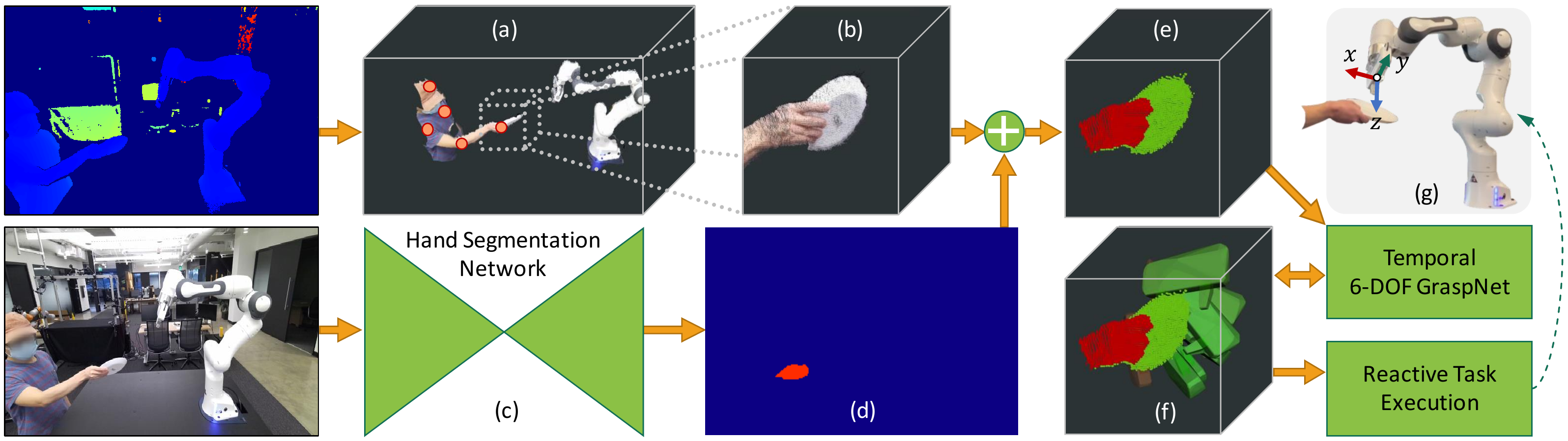}
    \caption{An overview of the proposed handover system. 
    Given RGBD images captured by an external Azure Kinect RGBD camera and the body tracking as input, the system first crops a point cloud that only contains the hand and the object. It then predicts a hand mask based on the RGB image. 
    Each point in the cropped point cloud can be labeled as hand or object using the hand mask. 6-DOF GraspNet is used to generate grasps or refine grasps from the last time step, conditioned on the object point cloud. The system removes grasps colliding with the hand point cloud and finally selects the best grasp and executes the handover through reactive task planning. 
    }
    \label{fig:overview}
\end{figure*}

In this paper, we present a system that can take arbitrary objects that are handed by a human to a robot.
Our approach takes advantage of the tremendous progress made on robot perception and grasping of arbitrary objects~\cite{mousavian2019graspnet,murali2020clutteredgrasping} while handling unique challenges of handovers. 
Unlike grasping static objects, human-to-robot handovers: (1) may involve the object being moved by the human, (2) involve objects that are not on an easily identified surface and partially occluded by human fingers, (3) are constrained in approach directions by the human's pose, especially their hand which is in contact with the object, and (4) require smooth robot motions that are intuitive and feel safe for the human.
To that end, we present a vision-based system that enables reactive human-to-robot handovers for unknown objects through real-time temporally consistent grasp generation and a closed-loop motion planning algorithm that ensures the reactivity and smoothness of the robot motion (Fig.~\ref{fig:overview}).

We demonstrate the generalization of our approach with (1) tests with a large benchmark of 26 household objects (Fig.~\ref{fig:teaser} and~\ref{fig:household-objects}) selected from Choi et al.'s study~\cite{choi2009list}, (2) a deeper systematic evaluation with three dramatically different objects examining handovers from different positions and orientations, as well as movement of the object after the robot starts moving, and 
(3) a user study ($N=6$) with the household objects examining handovers from different users.

%% file: 3_related_work.tex
Human-robot handovers have gained growing interest from robotics researchers over the past decade.
A recent survey~\cite{ortenzi2020object} summarizes the research progress made on different robotic capabilities that enable handovers, including  communication of intent, grasp planning, perception of the human, selection of the handover configuration, motion planning, grip force modulation for transfer, and error handling. Among the surveyed papers that involved real human-robot interactions, the majority focused on robot-to-human (R2H) handovers, while 11 involved human-to-robot (H2R) handovers. Most papers (26 out of 38) involved a single object being handed over; only one paper had more than six objects (i.e., Rosenberger et al. with 13 objects \cite{rosenberger2020object}). Hence, despite the great progress made in the field, there is a gap in generalization to larger sets of objects, or ultimately, arbitrary objects.

A simple yet practical approach for enabling H2R handovers is to keep the robot stationary and require the human to place the object in the robot's gripper \cite{edsinger2007human, aleotti2012comfortable}. 
While this approach can work in some situations, it places the burden of the handover on the human and is not applicable in assistive scenarios where the human motion might be restricted.
Instead, recent research aims to make H2R handovers reactive to the human and make the robot contribute to the transfer of the object. An important capability for enabling such H2R handovers is the recognition of the human's intent to hand an object to the robot.
Pan et al. address this problem with intent recognition algorithms trained with skeleton tracking data obtained in human-human handovers \cite{pan2017automated}.
Others used wearable sensing on the human for the same purpose \cite{wang2018controlling}.
Another challenge is to control the robot motion to take or receive the object from the human, which can be learned from human-human handover data \cite{yamane2013synthesizing,vogt2018one} or use control strategies like dynamic motion primitives \cite{maeda2017probabilistic,prada2014implementation}.
Several user studies investigated different aspects of H2R handovers with novice users in different scenarios \cite{koene2014experimental,koene2014relative,yang2020human}.
A benchmark was proposed in~\cite{sanchez2020benchmark} for H2R handovers of four different cups with unknown fillings to enable comparisons among different algorithms. 
However, these works have used a small set of objects, fixed or limited handover configurations, or both.

The core challenges we address in this paper are the perception of an unknown object being handed over, and the planning of a motion to grasp this unknown object.
Marturi et al. investigated the grasping of moving objects to enable human-to-robot handovers \cite{marturi2019dynamic}.
Our prior work contributed human grasp categorization algorithms to inform the choice of robot grasps for taking the object from the hand \cite{yang2020human}.
Konstantinova et al. aim to address the challenge of handing arbitrary objects, based on a method that relies only on wrist force sensors (i.e., no vision or tactile information), but still requires the person to bring the object in contact with the robot gripper \cite{konstantinova2017autonomous}.

Most closely related to this paper is recent work by Rosenberger et al.~\cite{rosenberger2020object}, who developed a method for grasping objects that can be recognized by the robot from a human giver's hand.
Their object detector is based on the YOLO V3 object detector~\cite{redmon2018yolov3}  trained on 80 object categories from the COCO \cite{lin2014microsoft} dataset and they generate grasps using a modified GG-CNN \cite{morrison2018closing}. Their approach was evaluated with 13 diverse objects handed in different configurations.
Our work has three key differences: 
(1) our system does not require objects to be in any pre-trained object dataset, which is a bottleneck of~\cite{rosenberger2020object};
(2) our system generates temporally consistent 6-DOF grasps using 6-DOF GraspNet~\cite{mousavian2019graspnet} instead of the planer grasps generated by GG-CNN \cite{morrison2018closing}, and this allows the users to handover objects in more unconstrained ways; 
(3) our system is closed-loop and the grasp selection is refined over time to enable reactivity by taking advantage of the robust tracking of the segmented hand and object.

%% file: 4_perception.tex
In this section, we present the segmentation module which enables accurate and real-time hand and object segmentation, 
and describe how we extend 6-DOF GraspNet~\cite{mousavian2019graspnet} with temporal refinement to generate temporally consistent and collision-free grasps given the segmented hand point cloud and object point cloud.

\subsection{Hand and Object Segmentation}\label{section:subsection:handobj}

To overcome the limitation of the object detector trained on a fixed set of objects as used in~\cite{rosenberger2020object}, one can use objectness or image saliency~\cite{alexe2012measuring, borji2015salient} for general object segmentation. 
These methods are usually trained on general objects (\eg, \textit{persons} and \textit{buildings}) with weak supervision, therefore they are incapable of generating accurate segmentation for unseen objects in hand, which would cause noisy grasp generation and unsafe motion planning.

In this paper, instead of directly segmenting the object in hand, we segment the hand given an RGB image, which is a more well-defined task compared with general unseen object segmentation, and then further infer the object segmentation with the depth information. 

\subsubsection{Hand Segmentation Model}
We trained a full convolution network for hand segmentation given an RGB image. 
Specifically, we use the Feature Pyramid Network~\cite{lin2017feature} based on ResNet-50~\cite{he2016deep} pretrained on ImageNet~\cite{russakovsky2015imagenet} as the backbone, and then stack four convolutions followed by upsampling operations to recover the feature map with the original resolution.  
Finally we generate a binary segmentation mask indicating whether a pixel is \textit{hand} or \textit{background}. 

\subsubsection{Inference}
At inference time, we track the user's hand by using the Azure Kinect Body Tracking SDK (Fig.~\ref{fig:overview}(a)), and then crop the point cloud centered at the user's palm center within a certain radius (20 cm in experiments), so that the cropped point cloud would cover both hand and the object in hand (Fig.~\ref{fig:overview}(b)). 
Then we project the points into the camera frame, and determine the hand point cloud $X_h$ and the object point cloud $X_o$ 
by the estimated hand mask (Fig.~\ref{fig:overview}(d)): the point is labeled as \textit{hand} if the corresponding pixel on the segmentation mask is \textit{hand}, otherwise it is labeled as \textit{object} (Fig.~\ref{fig:overview}(e)). 
Some example segmentation results are illustrated in Fig.~\ref{fig:hand-segmentation-dataset} (bottom) by projecting the hand and object point cloud into the camera frame.

\subsubsection{Network Training}
\begin{figure}[bt]
    \centering
    \includegraphics[width=1\linewidth]{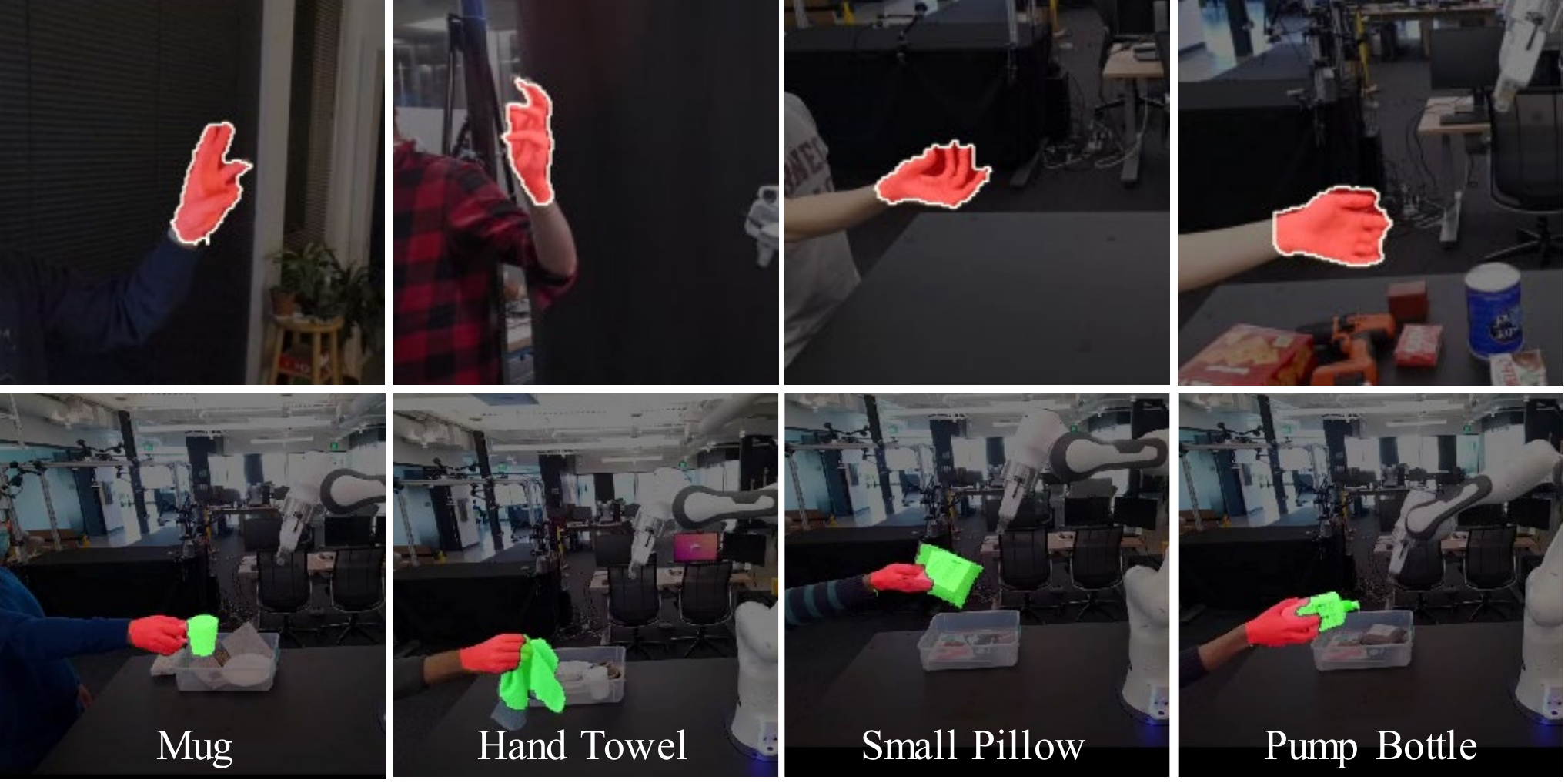}

    \caption{Top: the ground-truth hand masks (red) for training, which are generated on-the-fly without human annotations. 
    Bottom: the predicted hand (red) and object (green) segmentation masks. 
    Best viewed in color.}
    \label{fig:hand-segmentation-dataset}
\end{figure}

To facilitate the training, we collected a dataset with RGB images and the corresponding ground truth hand segmentation masks. 
Instead of recruiting costly human annotators, we reuse the idea of cropping the hand from the point cloud given hand tracking with a bare hand. The mask could be generated by projecting the hand point cloud into the camera frame. We further use the close morphological operation to remove the holes on the masks. 
Some examples are shown in Fig.~\ref{fig:hand-segmentation-dataset} (top). 
Though the masks generated in this way are less precise compared with human annotations, it enables us to collect a huge amount of data at a lower cost.
Furthermore, the segmentation model can be easily fine tuned to adapt to unseen environments, skin tones, or clothing by collecting more training data without costly manual annotations.

The segmentation model achieves $99.7\%$ Pixel Accuracy, $99.5\%$ fwIoU (Frequency Weighted Intersection over Union) and $68.0\%$ mean IoU on a test set collected in the same way.

\subsection{Temporally Consistent 6-DOF GraspNet}\label{subsection:graspnet}

6-DOF GraspNet~\cite{mousavian2019graspnet} is a learning-based framework which can generate a set of diverse 6-DOF grasps $G$ given the partial point cloud of an object $X_{o}$. 
Each grasp $g\in G$ is represented by $(R, T) \in SE(3)$ where $R \in SO(3)$ and $T \in \mathbb{R}^3$ are the rotation and translation of grasp $g$. 
GraspNet consists of two PointNet++~\cite{qi2017pointnet++} based networks: a \textit{Grasp Sampler} $g=f_S(z, X_o)$ to map latent variables $z\sim \mathcal{N}(0, 1)$ to a set of grasps for the object, and a \textit{Grasp Evaluator} $f_E(g, X_o)$ to assess the quality of a grasp by a score ranging from 0 (bad grasp) to 1 (good grasp). Both networks are conditioned on the point cloud of the object $X_o$.

We chose GraspNet as the grasp planner for our handover system of unseen objects since it is model-free and very efficient. 
The original GraspNet, however, is developed for grasping static objects. The grasps are generated beforehand and then the best one is chosen and executed. 
However, during a handover, humans cannot stay perfectly still, and may move or adjust their hands. Therefore we need to update the grasps over time. 
Generating grasps independently for each time step cannot guarantee the temporal consistency of grasps, which may lead to unstable robot motions.

\begin{figure}[bt]
    \centering
    \includegraphics[width=0.24\linewidth]{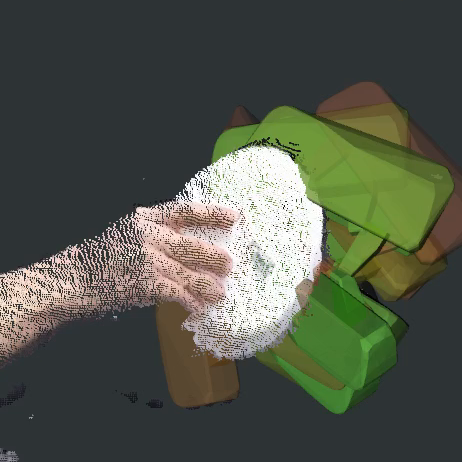}
    \includegraphics[width=0.24\linewidth]{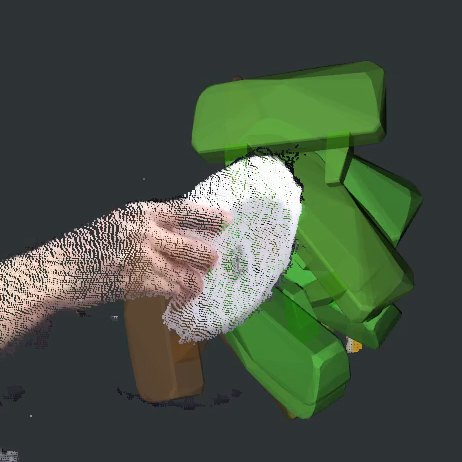}
    \includegraphics[width=0.24\linewidth]{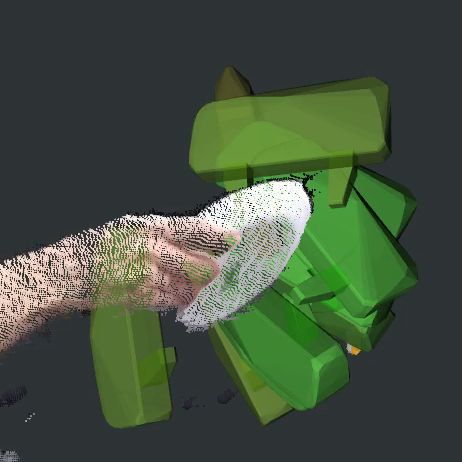}
    \includegraphics[width=0.24\linewidth]{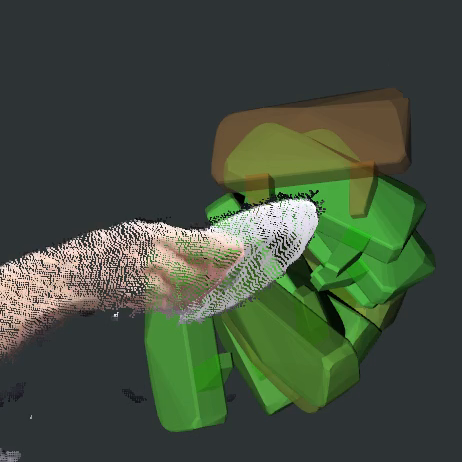}
    \caption{We generate consistent grasps over time when the object moves by the sampling-based refinement algorithm described in Sec.~\ref{subsection:graspnet}. Green denotes grasps with higher confidence and red denotes grasps with lower confidence. 
    }
    \label{fig:temporal-graspnet}
\end{figure}

To ensure the temporal consistency of the generated grasps, we extend the idea of grasp refinement in one static frame~\cite{mousavian2019graspnet} to consecutive frames overtime.  
Specifically, we refine the set of grasps $G_{t-1}$ generated from step $t-1$ based on Metropolis-Hasting sampling: 
for each grasp $g_{t-1} \in G_{t-1}$, we sample a perturbed grasp $g'$ as $g' = g_{t-1} + \Delta g$,
where $\Delta g$ is a small perturbation represented by $(\Delta R, \Delta T) \in SE(3)$. 
In practice, we set $\Delta R$ as an identity matrix and sample $\Delta T$ uniformly between $[-2, 2]$ centimeters.  

Then we evaluate the qualities of both grasps by the Grasp Evaluator $f_E$, and accept the perturbed grasps with the following probability: 
\begin{align}\label{eq:mh-ratio}
    r = \min \left(f_E(g', X_o)/f_E(g_{t-1}, X_o), 1 \right),
\end{align}
which ensures that the perturbed grasps is accepted if it leads to better grasp quality. On the other hand, if the perturbed grasp leads to lower quality we will accept it based on the probability above. This ensures that the grasp refinement process will not be stuck in the local minima. 
After that, we remove the grasps that are colliding with the hand point cloud $X_h$. 
Fig.~\ref{fig:temporal-graspnet} demonstrates that the grasps generated in this way are stable and consistent over time by maintaining relatively constant positions \textit{w.r.t} the object.

\begin{figure*}[ht]
    \centering
    \includegraphics[width=0.8\textwidth]{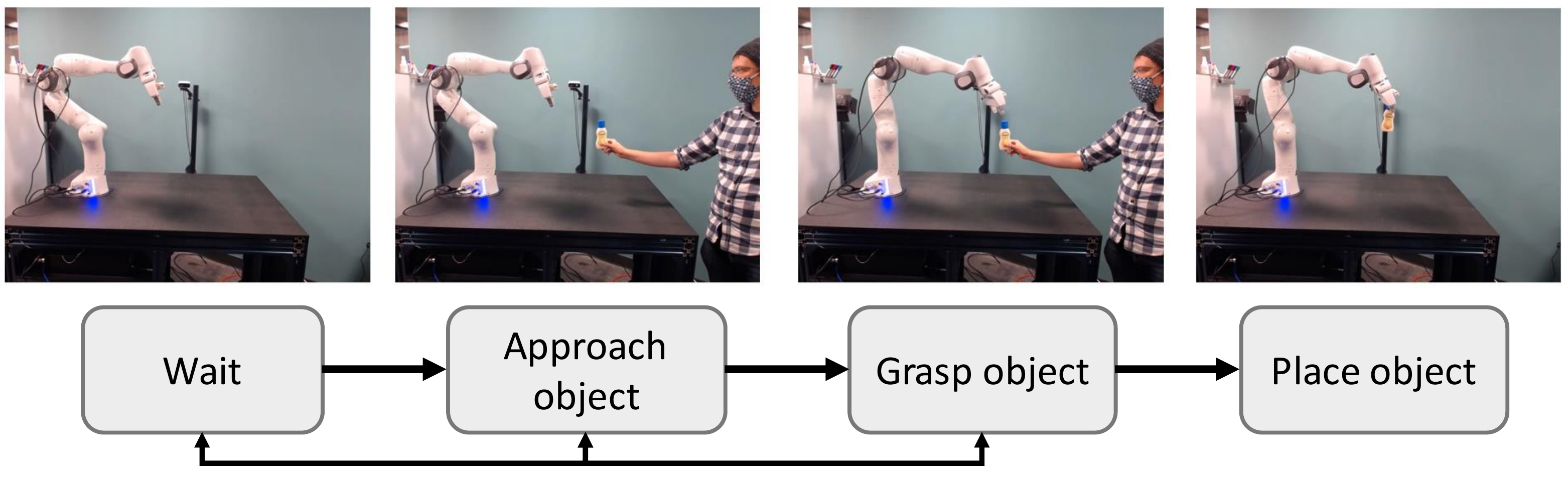}
    \caption{The stages in the task. The First three stages are waiting for the human, approaching the object, and taking the object; if any of these fail, or if the object shifts, the task model and planning system must react appropriately.}
    \label{fig:task-model}
\end{figure*}

\textbf{Special case. }
The above refinement process ensures the grasps are temporally consistent in most of the cases. When the hand motion is too fast, however, the point cloud of the object changes significantly, and the slightly perturbed grasps $g'$ would be mostly either colliding with or too far away from the object. 
Hence the ratio computed in Eq.~(\ref{eq:mh-ratio}) would be small, and the number of accepted grasps would decrease. 
In practice, we resample grasps when the number of grasps drops below a certain threshold. This ensures that our motion planner would have enough candidate grasps to generate reactive and natural robot motions.


%% file: 5_planning.tex
Our task model is based on planning with Robust Logical Dynamical Systems~\cite{paxton2019representing,kase2020transferable,yang2020human}: we compute a simple symbolic plan from a set of STRIPS-style high-level actions, with associated preconditions and effects.

We identified four main states the robot can be in, as shown in Fig.~\ref{fig:task-model}. We implement specific policies for each stage: 1) waiting at the home position for the human or object, 2) choosing a grasp and approaching to a standoff position aligned with the grasp position, 3) grasping, and 4) dropping the object. While waiting, the robot will move to its home position. The most important skill is 2), as this is when the robot is expected to follow the human's hand and choose which grasp to use. 

Given the robot's current observation of the world, we use a symbolic planner to sequence the available actions, and execute. At run time, we repeatedly check the conditions of every action in reverse order to determine which one to execute. If the robot has an object in hand, it will put it down (state 4); if the robot has chosen a grasp and reached its associated standoff position,
it will attempt to take the object (state 3). The robot attempts to approach the object (state 2) if the human's hand is above the table. If not, it will move back home and wait there (state 1).
This behavior constitutes 
a reactive hierarchical policy that can be used for reaching towards the human, as seen in prior work~\cite{yang2020human}.

Motions are implemented using Riemannian Motion Policies (RMPs)~\cite{ratliff2018riemannian}, which provide a basis for smooth, responsive human-robot interaction as in previous work~\cite{yang2020human}. For the purpose of safety and predictability, we also run a set of fast collision checks for each potential trajectory before following it with RMPs. Motion planning is done with the common RRT-Connect algorithm~\cite{kuffner2000rrt} when necessary, though we always attempt a straight-line path first.

The most important goal of the planning algorithm is grasp selection.
The set of grasps changes at every time step, as new grasp estimates come in from the temporally-consistent GraspNet. Due to sensor noise and small motions, these sets of grasps are not the same from one step to the next. As such, we propose a simple planning approach that generates trajectories that will be collision free and are kinematically feasible, while balancing the user's comfort and predictability of the resulting motion.

Given a set of candidate grasps, we first rotate every potential grasp $180^{\circ}$ around the grasp's Z-axis (Fig.~\ref{fig:overview}(g)), because our Franka Panda robot has a 2-finger gripper, and any grasp position can be additionally approached with the gripper upside down.
This gives us a larger pool of grasps with the associated grasp quality scores $s$ from the Grasp Evaluator $f_E$. We further compute a standoff ``approach'' pose $x_{appr}$ $10$cm back from the predicted grasp along its Z-axis, which is the configuration the robot will try to reach while tracking human's motions.
To ensure a firm grasp, we also have the robot push slightly into the object before closing the gripper, assuming that the human will adapt to the robot's motion. We move each grasp forward $5$cm along the $Z$-axis. 

We compute a score metric $C$ over the approach poses $x_{appr}$, which accounts for the distance between $x_{appr}$ and poses $x_{prev}$, the previous best grasp position, and $x_{home}$, the pose of the end effector at the home position:
\begin{align*}
    C = &w_{s} \min(s - s_{min}, 0)
    + w_{prev} d(x_{appr}, x_{prev}) \\
    &+ w_{home} d(x_{appr}, x_{home}),
\end{align*}
\noindent where $s_{min}$ the minimum acceptable grasp score and $w_{s}, w_{prev}, w_{home}$ are weights to bias grasp selection towards the home position, in order to maintain adaptability when the grasp predictions shift, while also maintaining temporal consistency.
$d(x_1, x_2)$ is a weighted quaternion distance function, as used in prior work~\cite{paxton2019prospection}. Given each pose $x = (p, q)$ where $p$ is the Cartesian position and $q$ is a unit quaternion describing orientation:
\[
    d(x_1, x_2)=  \|p_1 - p_2\|_2^2 + w_q (1 - \langle q_1, q_2 \rangle),
\]
\noindent where $w_q$ is a scaling parameter, trading off between position and orientation. 
In practice we set $w_s=1, w_q=0.1, w_{home}=5, w_{prev}=5$.

After computing the scores for all available grasps, we then loop in order and find inverse kinematics for each grasp, skipping it if no solution is found within a time limit. In practice we used Trac-IK for inverse kinematics~\cite{beeson2015trac}. Once an IK solution was found, we checked to make sure that (a) a collision-free joint-space path existed to this configuration, and (b) a collision-free Cartesian path existed from this standoff configuration to the final grasp position.
If the robot fails to find a grasp, it will ``track'' the object, staying slightly away from it in a collision-free configuration.

We update this grasp standoff position at roughly $10$Hz. 
When the robot arrives at this standoff position, 
it will initiate an open-loop grasp action: moving to the final grasp position and closing the gripper. If the gripper closes successfully around an object, the robot will move to state 4), drop the object, and terminate.

%% file: 6_experiments.tex
\begin{figure*}
    \centering
    \includegraphics[width=\textwidth]{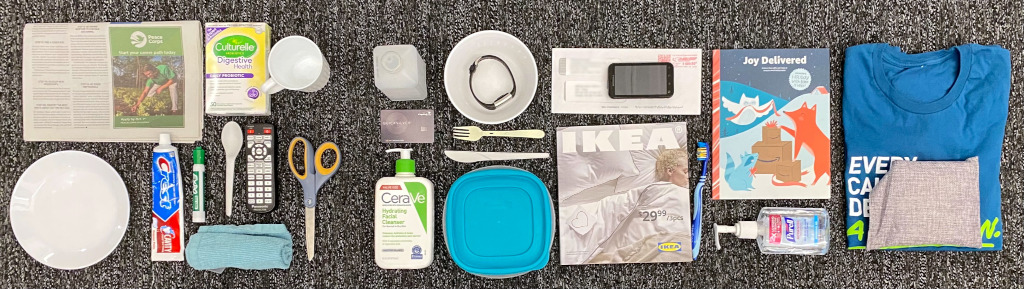}
    \caption{The 26 household objects we used for the user study. 
    We split it into subset \textit{Household-A} (left), which includes \textit{newspaper, medicine box, mug, plate, toothpaste, pen, spoon, remote, scissors} and \textit{hand towel}, for the constrained experiment, and subset \textit{Household-B} (right), which includes \textit{disposable bottle, dish bowl, credit card, fork, table knife, pump bottle, plastic container, letter, cell phone, book, toothbrush, magazine, hand sanitizer, small pillow} and \textit{T-shirt}, for the freeform experiment. 
    }
    \label{fig:household-objects}
\end{figure*}

\begin{table*}[bt]
    \centering
    \begin{tabular}{l cc cc cc cc}
    \toprule
         & \multicolumn{2}{c}{Banana} & \multicolumn{2}{c}{Bottle} & \multicolumn{2}{c}{Cereal Box} & \multicolumn{2}{c}{Overall} \\
         Grasp Type & Time (s) & Success (\%) & Time (s) & Success (\%) & Time (s) & Success (\%) & Time (s) & Success (\%) \\
    \midrule
        Flat & $11.87 \pm 0.33$ & 100\% & $8.46 \pm 0.30$ & 100\% & $10.42 \pm 1.50$ & 75\%
        & $10.27 \pm 1.65$ & 90\% \\
        45 degrees & $7.81 \pm 1.53$ & 100\% & $11.02 \pm 1.39$ & 75\% & $10.99 \pm 3.83$ & 75\%
        & $10.05 \pm 2.84$ & 82\% \\
        90 degrees & $7.94 \pm 0.58$ & 100\% & $18.59 \pm 5.77$ & 100\% & $15.89 \pm 3.73$ & 100\%
        & $14.14 \pm 6.02$ & 100\% \\
    \midrule
        Overall & $9.21 \pm 2.12$ & 100\% & $12.53 \pm 5.26$ & 90\% & $12.23 \pm 3.91$ & 82\% 
        & $11.39 \pm 4.29$ & 90\% \\
    \bottomrule
         Rotation & Time (s) & Success (\%) & Time (s) & Success (\%) & Time (s) & Success (\%) & Time (s) & Success (\%) \\
    \midrule
        0-45 degrees &  $ 16.34 \pm 2.92$ & 100\% & $10.86 \pm 1.55$ & 100\% & $8.23 \pm 1.11$ & 100\% &
        $11.81 \pm 3.93$ & 100\% \\
        0-90 degrees &  $16.47 \pm 8.49$ & 100\% & $18.40 \pm 6.89$ & 75\% & $12.89 \pm 6.15$ & 50\% & 
        $15.41 \pm 7.39$ & 69\% \\
    \bottomrule
    \end{tabular}
    \caption{Time and success rate of handovers. We completed handovers for three objects: a cereal box, a banana, and a mayonnaise bottle. Handovers were performed until the robot succeeded three times. \textit{Top}: objects were held at different orientations; different angles showed a different profile to the camera (running GraspNet), and made the motion planning problem easier or harder. \textit{Bottom}: we rotated the object partway through the motion, and our algorithm had to adapt to it.}
    \label{tab:grasp-type}
\end{table*}

We used an Azure Kinect RGBD camera mounted externally to capture the scene with $1280\times 720$ resolution and track the human body at $15$ FPS. The hand and object segmentation updated at $9$ FPS, and the grasps were generated and refined at $5$ FPS. We use DART~\cite{schmidt2014dart} to calibrate the robot arm and the camera. 
All the above modules shared a desktop computer with two NVIDIA RTX 2080ti GPUs. 
The task and motion planning ran on another computer.

\subsection{Handover Variations}\label{subsection:handover-variations}

To evaluate the robustness of our approach, we conducted experiments to see our system's performance in response to different types of handover and different objects with various sizes and shapes. 
Specifically, we looked at three different objects presented in three different orientations.

\begin{figure}[bt]
\centering
\includegraphics[width=0.32\columnwidth]{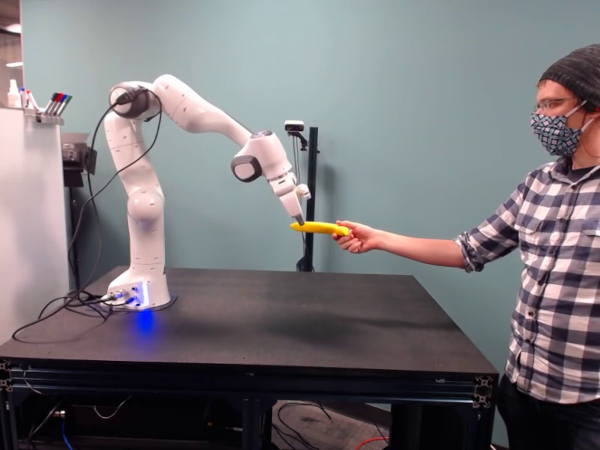}
\includegraphics[width=0.32\columnwidth]{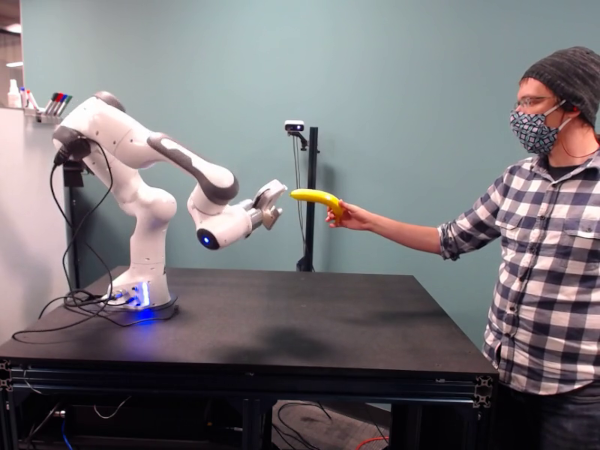}
\includegraphics[width=0.32\columnwidth]{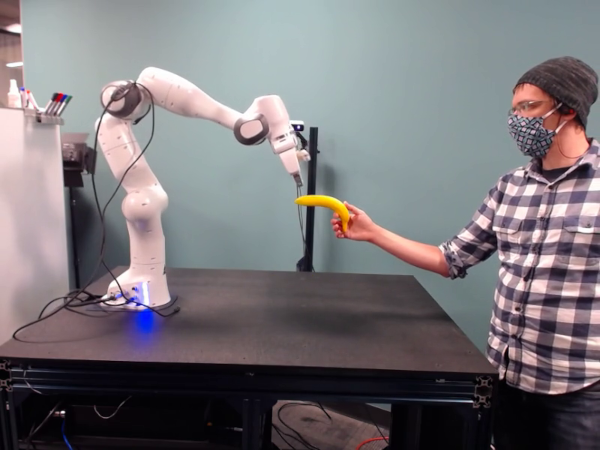}
\includegraphics[width=0.32\columnwidth]{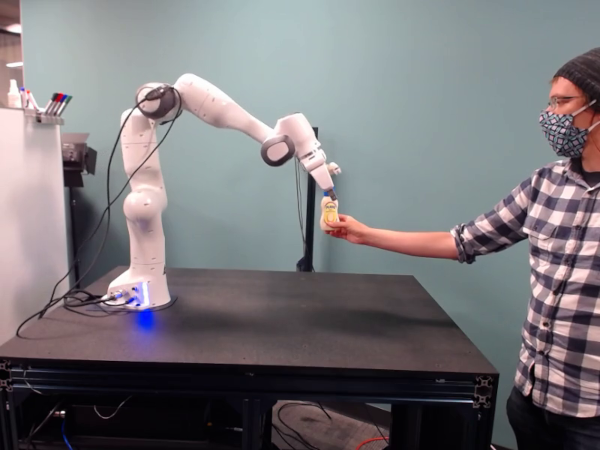}
\includegraphics[width=0.32\columnwidth]{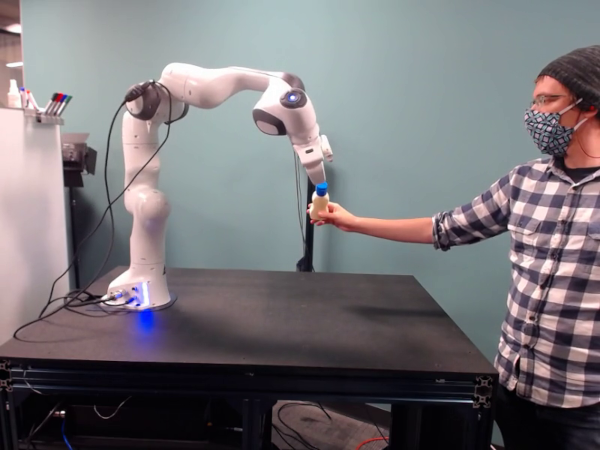}
\includegraphics[width=0.32\columnwidth]{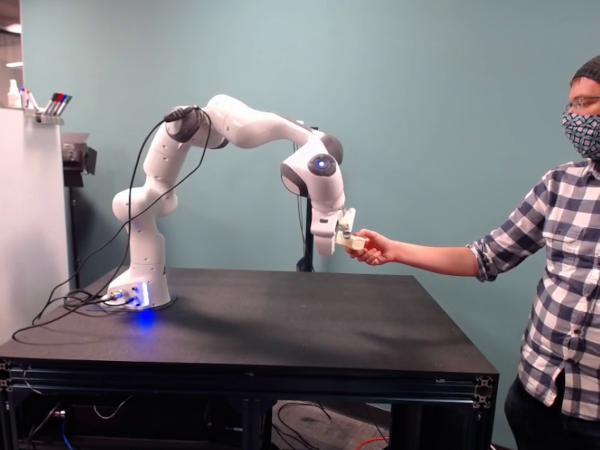}
\includegraphics[width=0.32\columnwidth]{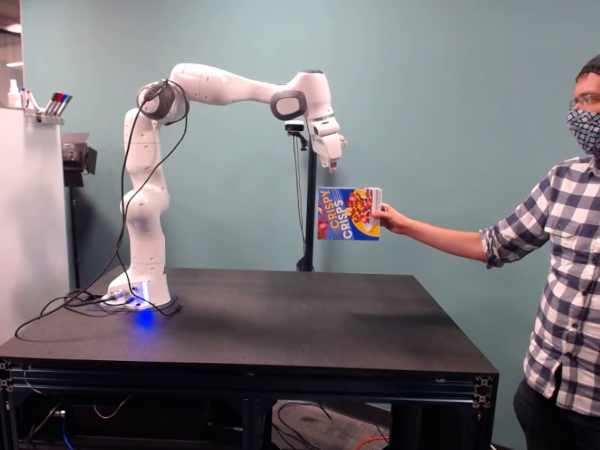}
\includegraphics[width=0.32\columnwidth]{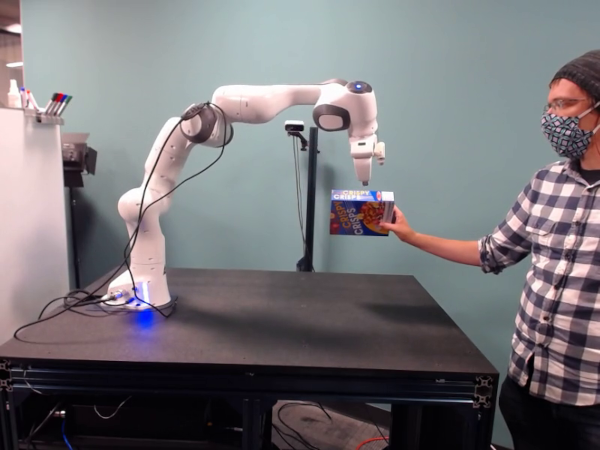}
\includegraphics[width=0.32\columnwidth]{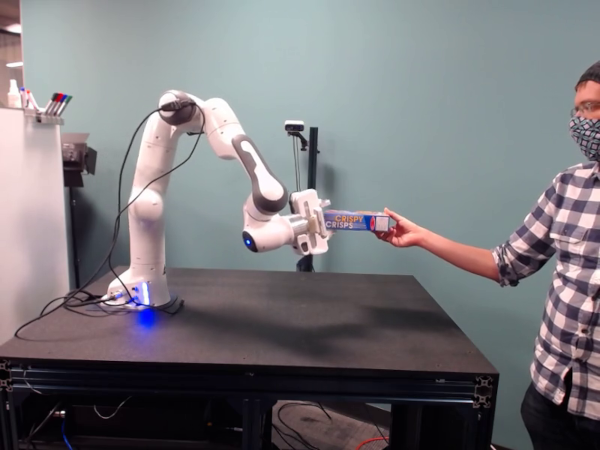}
\caption{Different objects and handover orientations used for the systematic evaluation. Three objects with different sizes and shapes were used: a banana, a toy bottle of mayonnaise, and a cereal box. The same user performed all experiments. 
All handovers were successful, though many took multiple tries for the robot to find a good grasp. 
}
\label{fig:systematic}
\end{figure}

As shown in Fig.~\ref{fig:systematic}, the objects were a plastic banana from the YCB object~\cite{calli2015ycb}, a toy bottle of mayonnaise, and a toy cereal box. Toy objects were from the HOPE dataset~\cite{tyree2019hope,tremblay2020indirect} and can be purchased online. The cereal box was the largest object, and the one that was not graspable from many different directions. 
However, it was also the easiest for the system to find grasps, since it was the easiest to see. The other two objects were much smaller, and required more precise grasps as a result.
We defined three human grasp types: 
one flat, one where the object is rotated at $45^{\circ}$, and one where it is rotated $90^{\circ}$, exposing a minimal profile to the camera in the case of the bottle or cereal box.

We measured time to successful handover, number of grasp attempts, and computed success rate for each one.
In addition, we recorded the number of attempts as the number of times the robot tries to approach and grasp the objects. 
Results are reported in Table~\ref{tab:grasp-type}. 
As expected, the cereal box was the hardest to grasp: it was the largest, and sometimes GraspNet failed to generate enough grasps  especially when it is flat with the large surface towards the camera.
During the bottle tests, we saw failures because of the bottle's shape: the system grasped in the air near the object because of the noisy grasps generated from the noisy point cloud near the object boundary. 
Overall, the system was able to grasp all three objects within three times.

\subsection{Handover Reactivity}
\label{sec:reactivity}

Next we examine the reactivity of our approach by changing the orientation of the object being handed after the robot starts moving.
Rosenberger et al.~\cite{rosenberger2020object} reported that most of their unsuccessful handover attempts were due to the object being moved after the robot had started to move, which our approach addressed through temporally consistent grasp generation in real time.
The bottom two rows in Table~\ref{tab:grasp-type} show the effect of changing rotation on handover performance. The object was initially in one of the orientations shown in Fig.~\ref{fig:systematic} and was rotated to another one once the robot started its movement.
As seen in Fig.~\ref{fig:systematic}, the three objects had very different profiles. We rotated the banana around its long axis, while the bottle was rotated around a shorter axis. 

Our system succeeded on taking the object in all cases within three attempts, demonstrating the reactivity of our approach.
Larger rotation changes resulted in slower handovers and more unsuccessful grasp attempts.


%% file: 7_user_study.tex
\begin{table*}
    \centering
	\setlength\tabcolsep{1.5pt}
    \begin{tabular}{llllllllllll}
        \toprule
          & Medicine Box & Newspaper & Plate & Mug & Remote & Toothpaste & Scissors & Towel & Pen & Spoon & Average \\
        \midrule
        Approach Time (s) & 
        $9.1\pm 3.5$ & $13.7\pm4.7$ &	$10.2 \pm 2.1$ &	$9.1 \pm 3.5$ & $9.7 \pm 0.7$ & $9.1 \pm 1.9$ &	$11.2 \pm 3.0$ & $13.8 \pm 3.8$ & $11.2 \pm 4.8$ & $10.4 \pm 1.9$ & $10.7\pm 3.6$ \\
        Number of Attempts & $1.3\pm 0.5$ & 1.0 & 1.0 & $1.8\pm 1.8$	& 1.0	& $1.2\pm 1.2$ & 	$1.2\pm 1.2$ &	1.0	& $2.0\pm 2.0$	& $1.5\pm 0.8$ & $1.3 \pm 0.3$ \\
        Success Rate & 75.0\% & 100\% & 100\% & 54.5\% & 100\% & 85.7\% & 85.7\% & 100\% &	50.0\% & 66.7\% & 81.8\% \\ 
         \bottomrule
    \end{tabular}
    \caption{Quantitative results from the user study with 6 users.
    }
    \label{tab:user_study}
\end{table*}

We conducted a user study with $6$ participants recruited from the lab\footnote{We were not able to recruit outside participants due to the pandemic.} to demonstrate our system's ability to handover a diverse set of objects. 

\subsection{Handovers with Diverse Household Objects}
To demonstrate the diversity of objects that our system can handover, we collected a diverse set of 26 household objects (Fig.~\ref{fig:household-objects}) from the list of objects for robotic retrieval tasks prioritized by people with ALS~\cite{choi2009list}. 
Some of handover trials on these objects are shown in Fig.~\ref{fig:teaser} and Fig.~\ref{fig:different-ways-handover}. 
Videos are available at \url{https://sites.google.com/nvidia.com/handovers-of-arbitrary-objects}.

\subsection{Study Setup and Protocol}

The user study involves two rounds of handovers from the participant to the robot. 
\textit{Round 1} involves a fixed set of 10 objects we selected from our household objects benchmark, which we refer to as \textit{Household-A} (Fig.~\ref{fig:household-objects} left). 
Objects were handed in a predefined order one by one. 
The users were told that the robot will start to move towards the object to take it from them once the object is not moving. 
\textit{Round 2} involved open-ended handovers where users were asked to pick any five objects they wanted from the rest of the household objects (referred to as \textit{Household-B}. See Fig.~\ref{fig:household-objects} right) and hand them over one-by-one to the robot. 
Optionally, users were also encouraged to handover other objects they could find in the lab or their personal belongings to the robot. 


After all the handovers were done, participants were asked to fill a questionnaire with Likert scale and open-ended questions and describe any problems they experienced. 

\subsection{Results}

\subsubsection{Round 1}
Fig.~\ref{fig:teaser} (top) shows the handover for all the 10 objects in the set \textit{Household-A} by participants in our study. The robot successfully grasped the object from all participants while avoiding the  contact with users' hands in all cases.
Table.~\ref{tab:user_study} reports performance metrics from Sec.~\ref{subsection:handover-variations} for handovers of \textit{Household-A} objects from user study participants.
Our system was able to take different objects from participants within a reasonable time and with a few trials.
Certain objects, such as \textit{Spoon}, \textit{Pen} and \textit{Mug}, took more trials for a successful grasp, due to the incomplete and noisy point cloud because of their smaller size, or the complexity of the object geometry. 



\subsubsection{Round 2}
\begin{figure}
    \centering
    \includegraphics[width=1\linewidth]{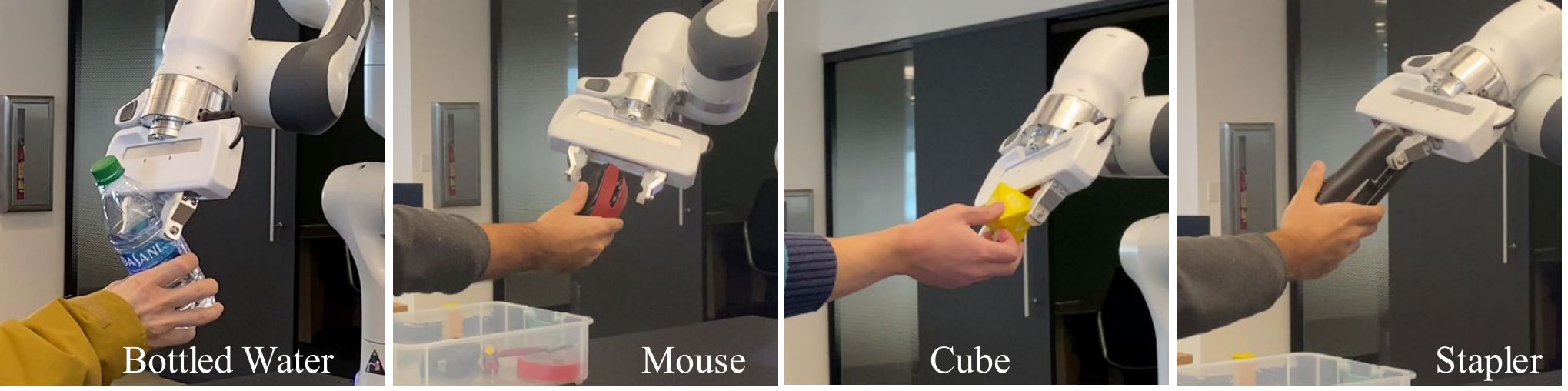}
    \caption{Handovers with objects that are not in the collected household objects.}
    \label{fig:other-objects}
\end{figure}

Handovers of objects from the set \textit{Household-B} by user study participants are shown in Fig.~\ref{fig:teaser} (bottom). Additionally, Fig.~\ref{fig:other-objects} shows handovers of objects outside of our benchmarks chosen by our participants. 
While we did not assess performance for these handovers, they  demonstrate our system's ability to handover diverse and truly arbitrary objects from different  users.

\begin{figure}[ht]
    \centering
    
    \includegraphics[width=1\linewidth]{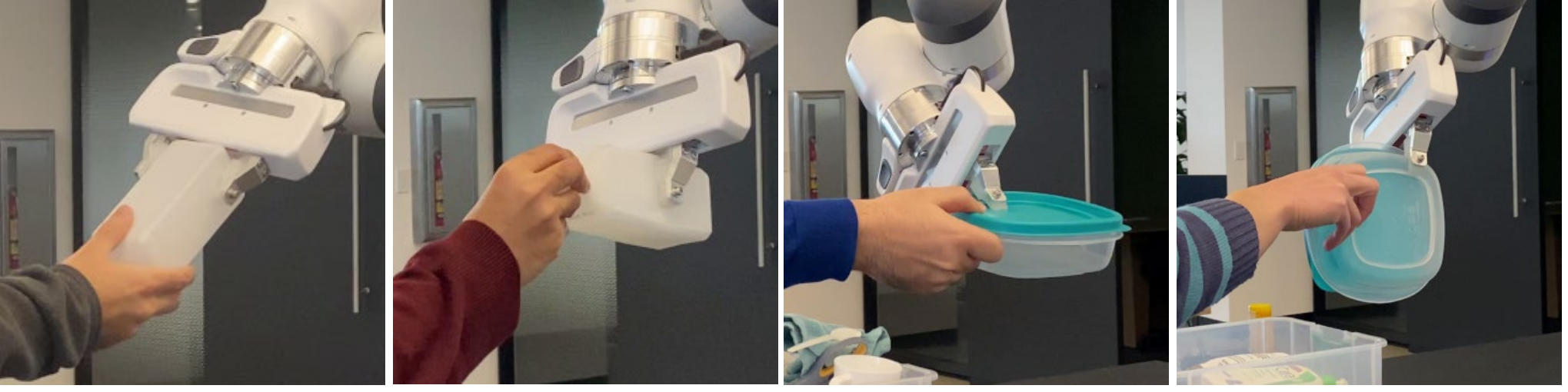}

    \caption{We observe that users have different preference on how to hold and handover an object. 
    Our system can adapt to these variance.}
    \label{fig:different-ways-handover}
\end{figure}

Fig.~\ref{fig:different-ways-handover} shows varying rotations in which different participants chose to hand objects. 
Our system was able to take the objects from humans in all these scenarios, which further validates the effectiveness of the system.




\subsection{Subjective Evaluation}

\begin{figure}[t]
    \centering
    \includegraphics[width=1\linewidth]{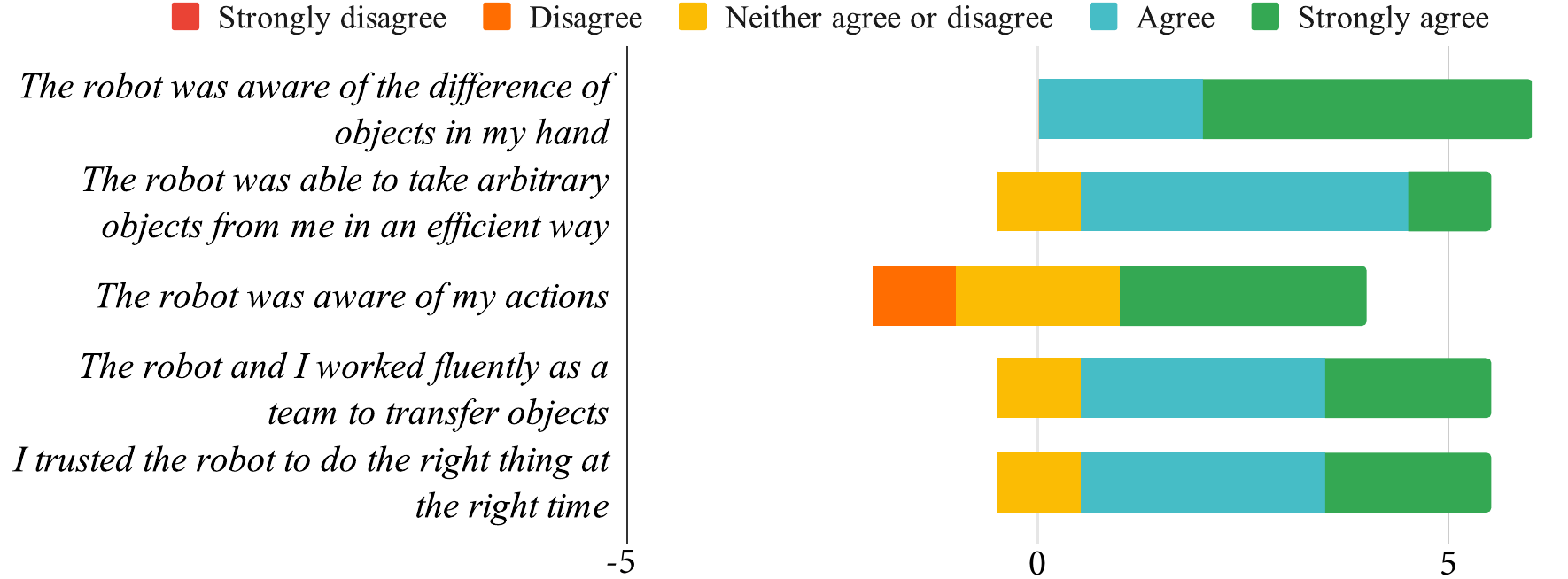}
    \caption{Participants' agreement with each statement in the questionnaire. 
    Horizontal axis denotes degrees of agreement with each statement.
    }
    \label{fig:handover-rating}
\end{figure}

Fig.~\ref{fig:handover-rating} presents responses to the questionnaire. 
Ratings in all dimensions were around 4 or above, with small variance across participants.
Participants thought that the robot was aware of the difference between objects, commenting that it was ``\textit{able to adapt its planned motion depending on the object in [their] hand}" and ``\textit{seemed to adjust to very different object geometries quite well.}"
Some noted the robot ``\textit{adapted the approach direction for objects that were highly asymmetric}" and one mentioned ``\textit{it was working properly for both deformable and rigid objects}".
Participants sense of safety was also confirmed by their comments, e.g., ``\textit{the robot is not going to pinch my hand}" and ``\textit{I've not felt any danger or potential harm to myself at any point during the test}."
Participants also noticed the reactivity of the handovers. One participant said
``\textit{I held the objects in different orientations, and often made adjustments in the middle of the robot's motion. The robot adjusted to these variations very fluidly}"; another said ``\textit{The robot adjusted to my motions}."

Participants thought the system could be improved in a few ways: one said ``\textit{the fluency can still be improved}" and another remarked that ``\textit{the robot is not very sharp in responding to [their] motion}". Others suggested the robot could ``move faster" and ``respond and adapt to [their] motion more efficiently".
One bottleneck in speeding up the robot motion as it continuously updates its target grasp on the tracked object is the high computation cost of inverse kinematics; which could be optimized through parallelism.
However, keeping the speed relatively low can be beneficial given its tradeoff with safety.

\subsection{Failure Cases}
Handover failures were due to one of three reasons: 1) the depth information was missing on dark surfaces due to excess absorption of light; 2) the object was recognized as the hand due to the failure of the segmentation network, which could be addressed by updating the segmentation model with new data 
from a wider variety of users;
3) the noise in the cropped object point cloud due to other nearby objects (\eg, objects in the bin). 
While the first two errors led to the robot not attempting to take the object as it could not compute a grasp, the last error led to grasps that missed the object.

%% file: 8_discussion.tex
We present a human-to-robot handover system that is generalizable to diverse unknown objects using accurate hand and object segmentation and temporally consistent grasp generation. 
There are no hard constraints on how naive users might present an object to the robot, so long as they hold it in a way that is graspable by the robot. 
The system can adapt to user preferences, is reactive to a user's movements, and generates grasps and motions that are smooth and safe.

%% file: appendix.tex
\onecolumn
\appendix
\subsection{System Setup}

\begin{figure}[ht]
    \centering
    \includegraphics[width=\linewidth]{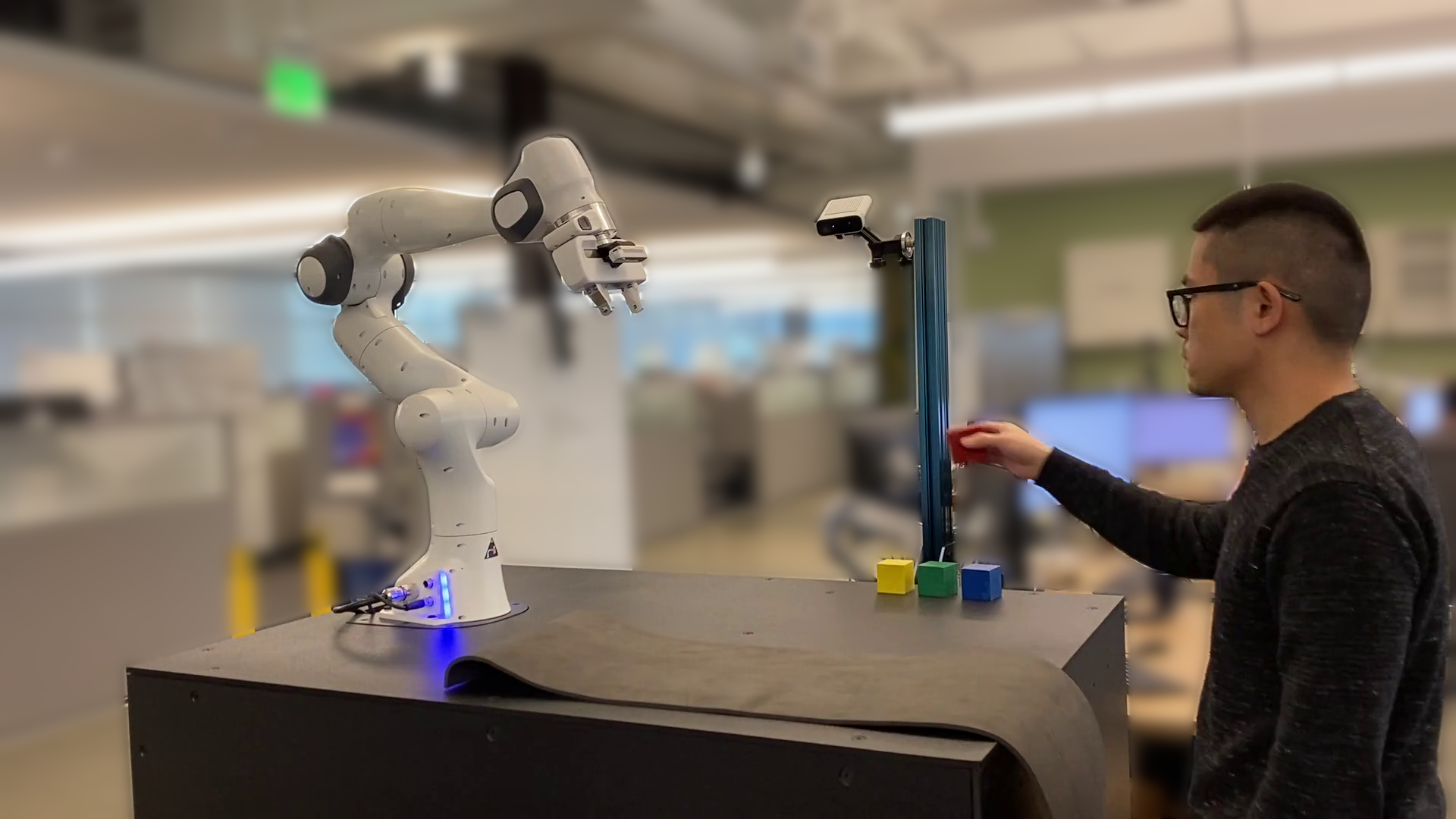}
    \caption{Our system consists of a Franka-Emika Panda robots mounted on a table, and an Azure Kinect RGBD camera mounted next to the robot. Three computers with three NVIDIA RTX 2080 Ti GPUs were used for perception, planning, and control.}
    \label{fig:system-setup}
\end{figure}

\begin{figure}
    \centering
    \includegraphics[width=\linewidth]{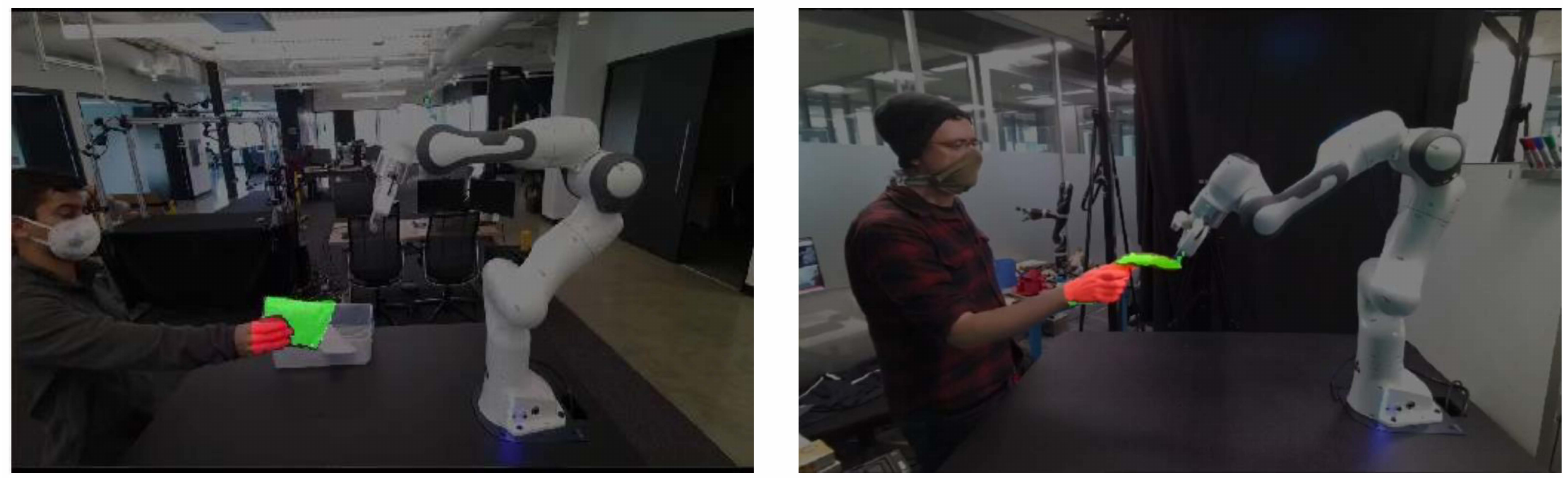}
    \caption{We conducted experiments with two Franka-Emika Panda robots mounted on identical tables in
two different locations.}
    \label{fig:two_locations}
\end{figure}

As shown in Fig.~\ref{fig:system-setup}, we used one Franka-Emika Panda robots mounted on a table for experiments. An Azure Kinect RGBD camera is mounted next to the robot. The camera was set to the wide field-of-view (WFOV) depth mode, and streamed RGBD images with 720P resolution and the human body tracking at 15 fps. Three desktop computers were used: one realtime computer for the control of the Franka, one for reactive task planning, and one for hand and object segmentation as well as grasp generation. We used DART~\cite{schmidt2014dart} to calibrate the camera to the robot. The hand-object segmentation, grasp generation, and DART were distributed on two NVIDIA RTX 2080ti GPUs. We duplicated this same setting and conducted experiments in two different locations, as shown in Fig.~\ref{fig:two_locations}.

\clearpage
\subsection{Ranking of the Object Difficulty Level}
\begin{figure}
    \centering
    \includegraphics[width=0.7\linewidth]{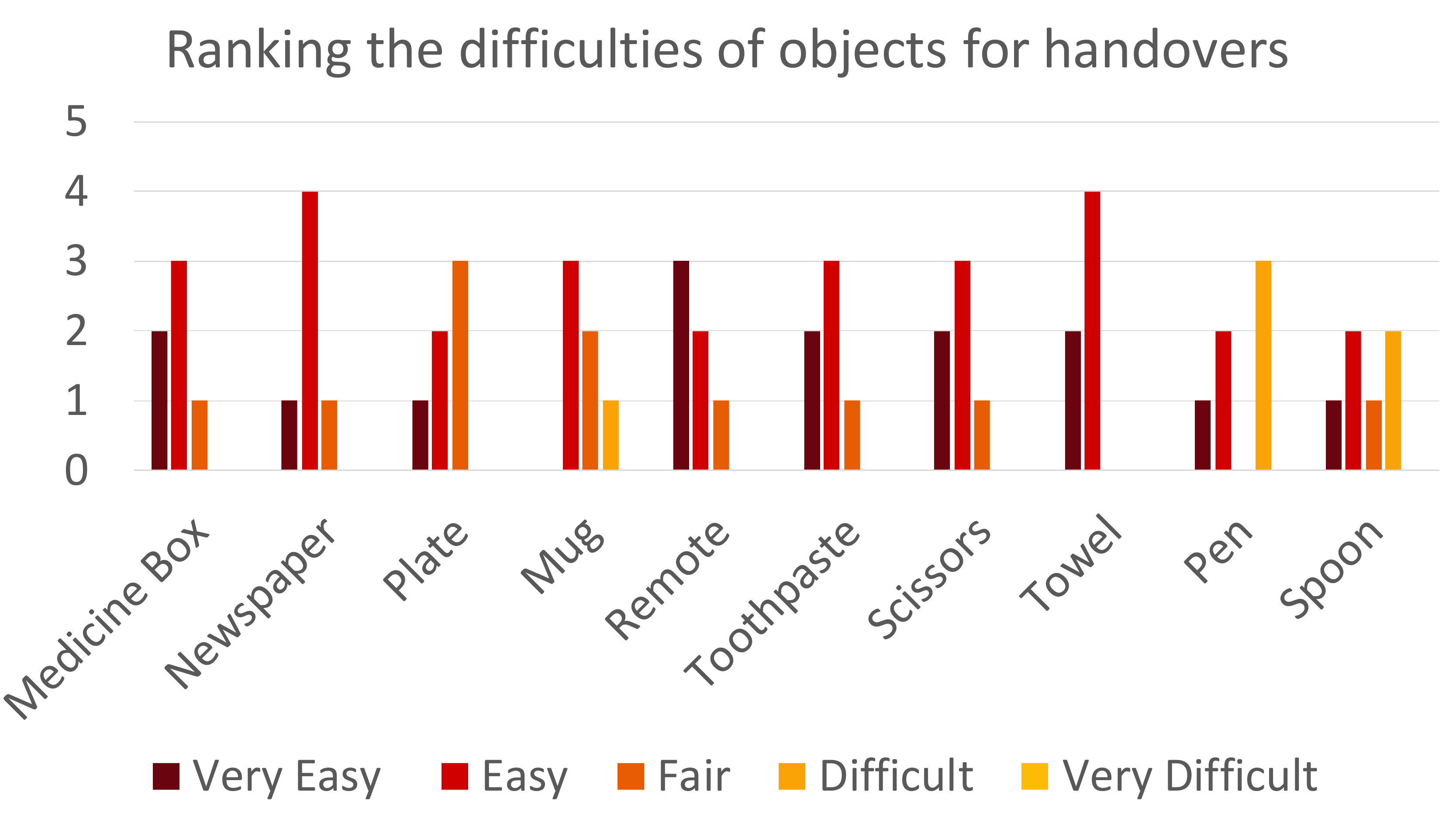}
    \caption{We asked users to rank the difficulty levels of objects from ``very easy" to ``very difficult" based on their experience.}
    \label{fig:object_difficulty}
\end{figure}

After the user study, we asked the users to rank the difficulty levels of \textit{Household-A} objects from ``very easy" to ``very difficult" based on their experience. The ranking is reported in Fig.~\ref{fig:object_difficulty}, which is consistent with the results in Table.~\ref{tab:user_study}. 

